\documentclass[11pt,a4paper]{article}
\usepackage[hyperref]{acl2020}
\usepackage{times}
\usepackage{latexsym}

\usepackage{enumitem}
\usepackage{graphicx}
\usepackage{float}

\usepackage{amssymb}
\usepackage{multirow}

\usepackage{pbox}
\usepackage{amsmath}

\usepackage{url}

\newcommand{\fontmed}{\fontsize{8.5pt}{8.5pt}\selectfont}

\usepackage[normalem]{ulem}
\useunder{\uline}{\ul}{}

\usepackage{microtype}

\aclfinalcopy 
\def\aclpaperid{1258} 

\title{Knowledge Fusion and Semantic Knowledge Ranking for Open Domain Question Answering}

\author{Pratyay Banerjee  \and Chitta Baral 
\\ Department of Computer Science, Arizona State University
\\ \texttt{pbanerj6,chitta}@asu.edu
}

\date{}

\begin{document}
\maketitle
\begin{abstract}

Open Domain Question Answering requires systems to retrieve external knowledge and perform multi-hop reasoning by composing knowledge spread over multiple sentences. In the recently introduced open domain question answering challenge datasets, QASC and OpenBookQA, we need to perform retrieval of facts and compose facts to correctly answer questions. In our work, we learn a semantic knowledge ranking model to re-rank knowledge retrieved through Lucene based information retrieval systems. We further propose a ``knowledge fusion model'' which leverages knowledge in BERT-based language models with externally retrieved knowledge and improves the knowledge understanding of the BERT-based language models. On both OpenBookQA and QASC datasets, the knowledge fusion model with semantically re-ranked knowledge outperforms previous attempts.  

\end{abstract}

\section{Introduction}

Open Domain Question Answering is one of the challenging Natural Language Question Answering tasks where systems need to retrieve external knowledge and perform multi-hop reasoning by understanding knowledge spread over multiple sentences.
In recent years, several open domain natural language question answering datasets and challenges have been proposed. These challenges try to replicate the human question answering setting where humans are asked to answer questions and are allowed to refer to books or other information sources available to them.
Datasets such as HotPotQA \cite{yang2018hotpotqa}, Natural Questions \cite{kwiatkowski2019natural}, MultiRC \cite{MultiRC2018}, ComplexWebQuestions \cite{talmor18compwebq} and WikiHop \cite{welbl2018constructing} require finding relevant knowledge and reasoning over multiple sentences. In these tasks, the systems are not constrained to any pre-determined knowledge bases. Both, the task of finding knowledge and reasoning over multiple sentences requires deep natural language understanding. The goal of these tasks is not to memorize the texts and facts, but to be able to understand and apply the knowledge to new and different situations \cite{jenkins1995open,Mihaylov2018CanAS}.

\begin{figure}[t]
\begin{center}
  \includegraphics[width=0.8\linewidth]{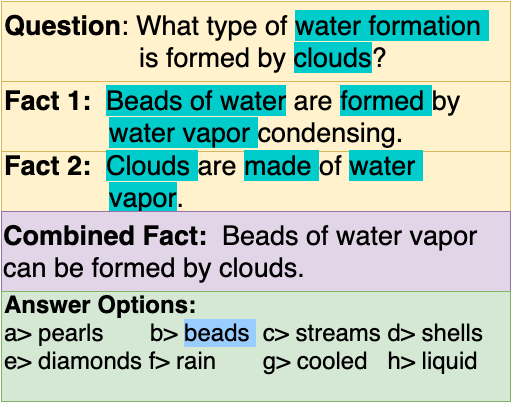}
\end{center}
   \caption{An example from the QASC Dataset. The highlighted words are the keywords present in the Question, Annotated Fact 1 and Annotated Fact 2. Annotated Facts 1 and 2 are present in the associated knowledge corpus. These are the gold knowledge facts that need to be retrieved and composed to answer the question. The correct answer is highlighted in blue.}
\label{fig:motivation}
\end{figure}

In our work, we focus on the datasets, OpenBookQA and QASC \cite{Mihaylov2018CanAS,khot2019qasc}. They differ from the above mentioned datasets in the following aspects. Special care is taken in OpenBookQA and QASC to avoid simple syntactic cues in questions, that allow decomposition into simpler queries \cite{khot2019qasc}.
It has been shown through human verification that answering questions in both of these datasets requires a composition of two or more facts.
OpenBookQA is accompanied by an open-book of facts, which contains partial knowledge to answer the questions. QASC provides a knowledge corpus of around 19 million science facts. 

In OpenBookQA, to answer questions systems need to combine a seed core fact from the provided open-book with an unknown number of additional facts and an unspecified source for retrieving such additional facts. Questions in QASC require retrieval of exactly two facts from the provided knowledge corpus. Moreover, it provides annotation for which facts are required, allowing the building of a supervised knowledge retrieval model.

There are several challenges we need to address while solving these datasets. The first challenge is the task of relevant knowledge retrieval. We use a simple Lucene based information retrieval system Elasticsearch \cite{gormley2015elasticsearch} to retrieve knowledge from an appropriate knowledge corpus. The task of retrieval is done for each answer options. Moreover, we require multiple steps during retrieval as it has been mentioned above, the questions need composition of knowledge in multiple sentences to correctly answer the questions. A multi-step information retrieval introduces a significant amount of noise. We address this challenge by learning a BERT-based Semantic Knowledge Ranking model. 

The second challenge we address is the creation of the dataset for Semantic Knowledge Ranking. In order to train our Semantic Knowledge Ranking model, we need knowledge facts with both positive and negative labels. For this task, QASC, OpenBookQA and SciTail \cite{scitail} provide positive labels, i.e, the knowledge facts which  are relevant to a given question-answer pair. We address the challenge of negative label creation using our automatic dataset preparation techniques. The design choices for the dataset preparation are demonstrably impactful on the downstream question answering task.

The third challenge we address is about knowledge composition and understanding. BERT-based language models possess considerable knowledge learned through their pre-trained language modelling tasks. The challenge of incorporating external domain specific knowledge, discriminative enough to answer questions is addressed by our knowledge fusion module over BERT-based language models. We identify the drawbacks in the current way systems use BERT-based language models, and augment them with our new knowledge fusion module.

Finally, we analyse both our knowledge ranking and question answering models to identify how different components contribute. We also analyze what noise our knowledge ranking model introduces, where our question answering model fails and why is it unable to answer such questions. Our analysis shows some of the drawbacks of using Attention-based language models and the necessity of improvements in specific components.


Our contributions are summarized below:
\begin{itemize}[noitemsep]
    \item We prepare a Semantic Knowledge Ranking dataset by using annotations from three different sources and with our dataset preparation techniques. We also evaluate two BERT-based ranking model on this task.
    \item We propose a new model to perform better knowledge composition and question answering with external knowledge.
    \item We analyze our knowledge ranking and knowledge composition models to better understand the failures to enable future improvements. 
    \item We achieve new state-of-the-art results in both OpenBookQA (+2.2\%) and QASC (+7.28\%).  
\end{itemize}

\begin{figure*}
  \includegraphics[width=\linewidth,height=7cm]{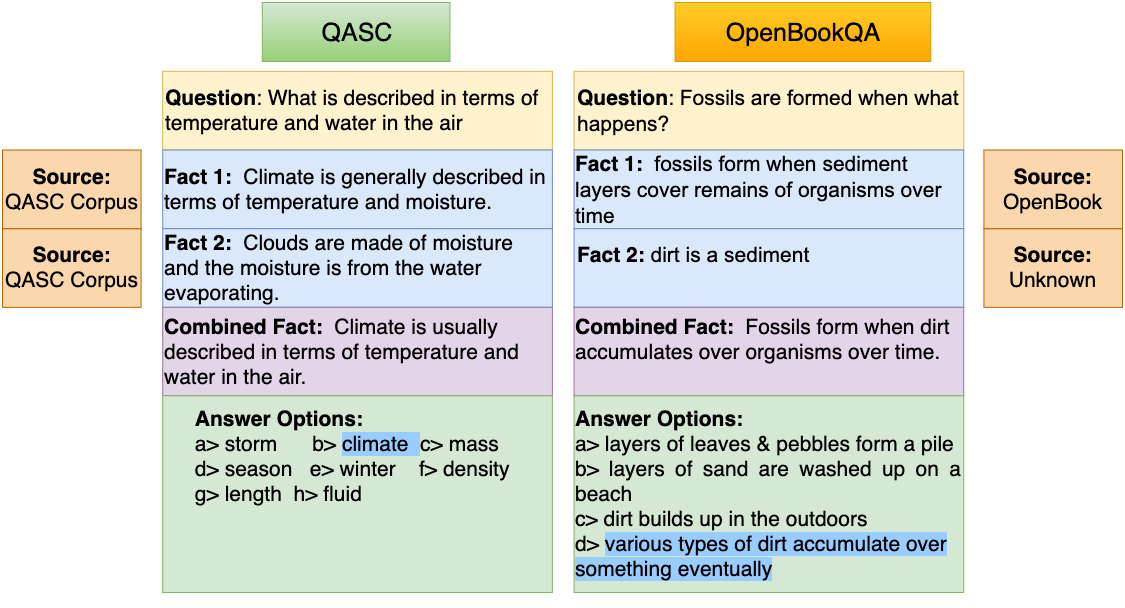}
  \caption{Examples of questions present in OpenBookQA and QASC. The source of facts for QASC is the available knowledge corpus. For OpenBookQA, the Fact 1 is present in the open-book, but the Fact 2 needs to be retrieved from an external source. Correct answers are highlighted in Blue.}
  \label{fig:example}
\end{figure*}

\section{Related Work}
\paragraph{Text based QA:}
Multiple datasets have been proposed in recent years for natural language question answering such as \cite{rajpurkar2016squad,joshi2017triviaqa,khashabi2018looking,richardson2013mctest,lai2017race,reddy2018coqa,choi2018quac,tafjord2018quarel,mitra2019declarative,sap2019socialiqa,bisk2019piqa}. There are several attempts to solve these challenges such as \cite{devlin2019bert,vaswani2017attention,seo2016bidirectional,liu2019roberta} etc. In our work we focus on QASC \cite{khot2019qasc} and OpenBookQA \cite{Mihaylov2018CanAS} datasets, which are multiple-choice question answering datasets which need multi-hop reasoning and knowledge retrieval from provided knowledge corpus.
\paragraph{External Knowledge:}
Closest to our work are the models which use external knowledge, such as in the work of \cite{khot2019s,khot2019qasc,pirtoaca2019answering,banerjee-etal-2019-careful,mitra2019exploring} where they use sentences as knowledge. We differ to these with our knowledge ranking and infusion models. Other models such as in the work of \cite{mihaylov2018knowledgeable,chen2018neural,yang2019enhancing,wang2019explicit} embed syntactic or semantic knowledge into knowledge enriched embeddings. They do not use additional sentences as knowledge. Sentences lack a well-defined structure and possess a larger number of variables. Semantic knowledge retrieval is also present in a different form in the above mentioned work. The task of neural explanation retrieval or ranking such as in \cite{jansen-ustalov-2019-textgraphs,banerjee-2019-asu} is similar to semantic knowledge retrieval. We differ from them in the way we model our tasks and our dataset preparation techniques.

\section{Datasets}
\paragraph{Open Book Question Answering:}
OpenBookQA is a multiple-choice school level science question answering dataset from AllenAI. There are four answer options for each question. The dataset provides an accompanying open-book of 1,324 science facts. There are a total of 4,957 questions in train and 500 questions in each validation and test set. The nuance of this dataset is that it needs knowledge retrieval and knowledge composition to answer the questions. It is specified that one of the facts is present in the accompanying open-book and we need additional knowledge sources to retrieve the other facts. 

\paragraph{Question Answering via Sentence Completion:}
QASC is another multiple-choice science question answering dataset from AllenAI. There are 9,980 8-way multiple-choice questions from elementary and middle school level science. There are 900 questions in the validation and 920 questions in the test set. It is different from OpenBookQA as it is accompanied by a knowledge corpus of 19 million science facts. Moreover to answer each question we need to combine exactly two facts, both of which are present in the knowledge corpus. QASC provides additional annotations of which two facts are required to answer the questions. These annotations enable us to create a semantic knowledge ranking dataset. Figure \ref{fig:example} shows an example from both of the datasets.

\section{Approach Overview}

Our approach includes the following modules: Multi-step Knowledge Retrieval, Semantic Knowledge Ranking Dataset Preparation, Semantic Knowledge Ranking and Knowledge Fusion based Question Answering.

In \textbf{Multi-step Knowledge Retrieval} we want to extract relevant facts from the knowledge corpus. We use heuristic-based query creation techniques. In the first step of the retrieval, we use the question $Q$ and the answer option $A_i$ to generate queries. In the second step, we use question $Q$, the answer option $A_i$ and $F_1$, the top ten facts retrieved and ranked from step one, to generate queries. 

In \textbf{Semantic Knowledge Ranking}, we re-rank our retrieved knowledge sentences using BERT-based language models. To train our models, we do \textbf{Semantic Knowledge Ranking Dataset Preparation} by using the provided annotations present in the QASC, OpenBookQA and SciTail datasets. But the annotations only provide positive labels, so we use our dataset preparation techniques to automatically generate negative labels.

We train a BERT-based language model for question answering using external knowledge and identify the gaps in knowledge understanding. We analyze the results and define a \textbf{Knowledge Fusion} module. The module is based on a simple intuition that the model should have the ability to compare against different answer options and the input should clearly distinguish between the knowledge which is ranked higher across all answers. 

\section{Multi-Step Knowledge Retrieval}
\paragraph{Knowledge Sources:}
We use the open-book from OpenBookQA, the knowledge corpus from QASC and the ARC knowledge Corpus \cite{clark2018think} as our knowledge sources. There are 1324 knowledge facts in the open-book from OpenBookQA. There are 19 million knowledge facts in the knowledge corpus of QASC. There are 1.7 billion knowledge facts in ARC knowledge corpus. We keep the knowledge facts as is for the open-book and QASC knowledge corpus. We preprocess the knowledge facts from ARC by removing non-English characters and remove punctuations that do not have an impact on sentence structure. We index the factual sentences into Elasticsearch, a Lucene based search engine. 

\paragraph{Step-1:}
In the first step of knowledge retrieval, we have a question $Q$ and the $ith$ answer options $A_i$. We generate the query using a simple heuristic of concatenating the question and answer, and removing stop-words but keeping the order of the words intact. We use stop-words from NLTK and standard python ``stop-words'' library. We query the Elasticsearch IR system and retrieve top-50 sentences. These sentences are denoted as $F_1$.

\paragraph{Step-2:}
The second step of knowledge retrieval is similar to the work of heuristic-based natural language abduction in \cite{banerjee-etal-2019-careful} and the 2-step retrieval mentioned in \cite{khot2019qasc}. For each question $Q$, answer option $A_i$ and top ten knowledge retrieved and semantically ranked $F_{1ij}$ we find the set of unique words present in $Q$, $A_i$ and $F_1ij$ using the following heuristics:
\begin{equation*}
    Qu_{ij} = ((Q \cup A_i) \cup F_{1ij}) \setminus ((Q \cup A_i) \cap F_{1ij}) 
\end{equation*}
where $Qu$ is the generated query, $ith$ answer option, and $j$th retrieved sentence from $F_{1i}$. These sentences are denoted as $F_2$.

\section{Semantic Knowledge Ranking}
\subsection{Task Definition}
We model the semantic knowledge ranking task as a binary sentence-pair classification task, i.e, given the question $Q$ and answer option $A_i$, we classify the corresponding retrieved knowledge $F_{kij}$ into two classes, irrelevant and relevant, where $k$ is the IR step, $i$ is the answer option number and $j$ is the retrieved sentence number. We rank the sentences using the class probabilities for the relevant class. 

Formally, we learn the following probability:
\begin{equation}
    Rel(F_{kij},Q,A_i) = P(F_{kij} \in G|Q,A_i)
\end{equation}
where $G$ is the set of relevant facts.  

We compare multiple task settings for Knowledge Ranking, such as a regression task similar to Semantic Textual Similarity. Though this task is more appropriate as a ranking task, it is harder to get correct and noiseless annotations for such a task using automatic techniques. 

\subsection{Dataset Preparation}
For the classification task, we require both relevant and irrelevant facts given a question and answer pair, which is simpler to achieve automatically. 

\paragraph{Positive Labels:} Questions in QASC are accompanied by two human-annotated knowledge facts, which when composed can be used to answer the questions. These annotated facts provide us the positive labels. We gather more positive labels from the OpenBookQA datasets. The OpenBookQA also has an additional resource where they provide which fact from the open-book is the most relevant. The final source of positive labels is the SciTail dataset \cite{scitail}. SciTail contains questions, the correct answer, and a sentence pair. The task in SciTail is natural language inference, i.e, does the fact entails or is neutral to the hypothesis (question-answer). The hypothesis is not a concatenated version of the question and an answer but a well-structured sentence. Unfortunately in our target application, we do not possess a well-structured sentence equivalent to our question-answer pairs. For positive labels, we select question-answer pair and the corresponding premise as the relevant facts from the samples annotated as ``entails''.

\paragraph{Negative Labels:} From SciTail we take all samples marked as ``neutral'' as an initial set of irrelevant facts. We gather further negative samples using the following simple heuristics. For all question-answer pairs from QASC and OpenBookQA we do an initial knowledge retrieval using the query generation as mentioned in step one of multi-step knowledge retrieval. From this set of retrieved facts we identify similar facts to the gold relevant facts using Spacy \cite{spacy2} document similarity and mark them as irrelevant, i.e, sentences retrieved using wrong answers similar to gold whose similarity is less than a threshold of $T$. $T$ is identified using manual evaluation of 100 samples. Spacy uses word vectors to compute document similarity. The similarity function used is cosine similarity. We also mark the knowledge retrieved using wrong answer options as irrelevant. Note that we never mark the gold annotated relevant facts as irrelevant for the wrong answer options, as we want our Semantic Knowledge Ranking model to rank these facts better even for the wrong answer options. 
Train and validation sets are balanced for both classes, with each having 145,200 and 16,134 samples respectively.

\subsection{Model Description}
We evaluate two BERT-based language models, BERT-large-cased \cite{devlin2019bert} and RoBERTa \cite{liu2019roberta} for the sentence-pair classification task.

We provide the concatenated question and answer pair as sentence A and the fact as sentence B. Let $S_A$ denote tokens from sentence A and $S_B$ denote tokens from sentence B, then the input to the BERT model is defined as $\{[CLS] S_{A_i}  [s]  S_{B_j}  [s] \}$. Here $[s]$ is the separator token. $[CLS]$ is the class token. We take the encoding of $[CLS]$ token from the final layer of BERT, which we pass through a feed-forward layer of dimension $H_{dim} \times 2$, where $H_{dim}$ is the dimensions of the encoding and a final softmax layer for getting probabilities for the two classes. We use cross-entropy loss between the predicted scores and the gold relevance labels. \footnote{More training details is in Supplementary Materials.}

\begin{equation*}
    logits = FF(\beta(CLS,S_A,S_B))
     \end{equation*} 
    \begin{equation*}
    score(F_{kij},Q,A_i) = softmax(logits)
 \end{equation*} 
$\beta$ is the BERT model, FF is the feedforward layer.


\begin{figure*}
  \includegraphics[width=\linewidth,height=6.8cm]{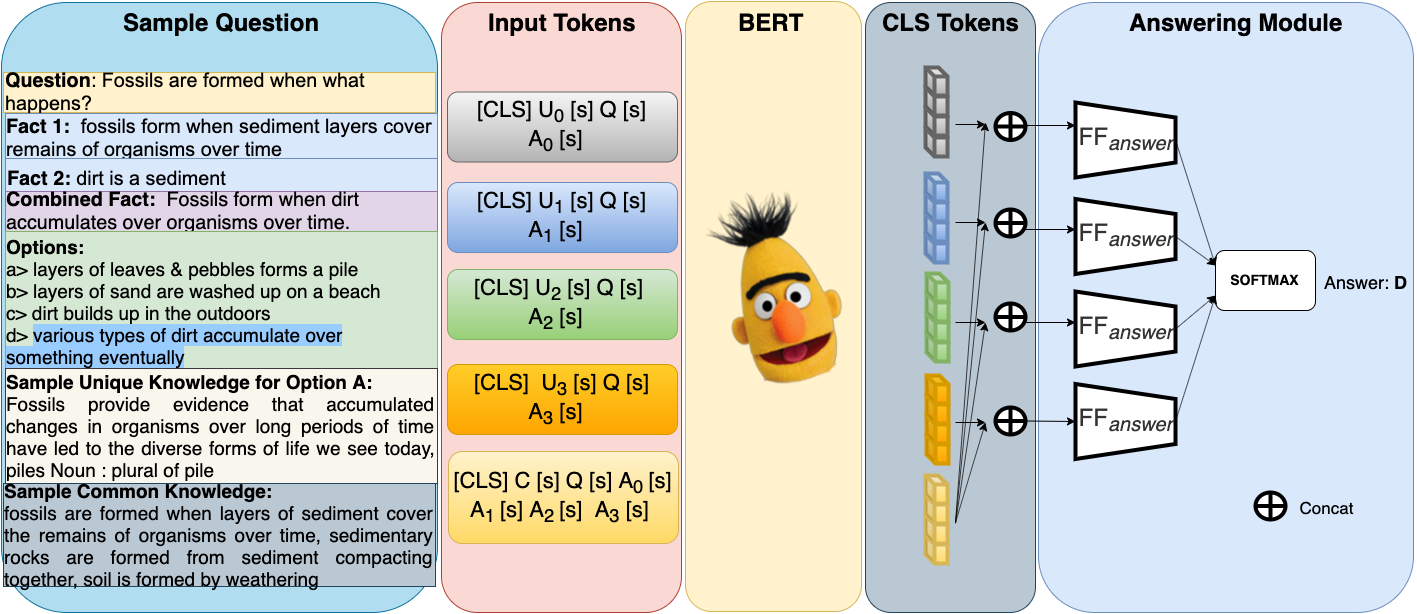}
  \caption{Our end-to-end model view with sample question, input description and knowledge fusion module. The correct answer is highlighted in blue. The sample knowledge are retrieved from OpenBookQA and our knowledge corpus, consisting of ARC Corpus, OpenBookQA open-book and QASC knowledge corpus. The sample knowledge are semantically ranked.}
  \label{fig:fusion}
\end{figure*}

\section{Knowledge Fusion Question Answering Model}
\paragraph{Overview of BERT-based QA Models:}
Let $Q$ define the set of tokens in the question, $A_i$ the set of tokens in the $i$th answer option and $K_i$, the retrieved knowledge. The current systems \cite{devlin2019bert,liu2019roberta,banerjee-etal-2019-careful,khot2019qasc} use the following way of defining the input to the BERT-based models. For an answer, they append the question tokens $Q$ to the knowledge tokens $K_i$. The total input is defined as $\{[CLS] \, K_i \, Q \, [SEP] \, A_i \, [SEP]\}$. This can be interpreted as an entailment model where, given the knowledge and question, we predict the entailment score of each answer. Each answer has a separate input. This way of creating input and modeling question answering has certain drawbacks.

Firstly, each knowledge retrieved is unique to the corresponding answers. BERT has multiple layers of stacked attention neural units, and attention with the corresponding knowledge enables the system to perform the question answering task. But the knowledge retrieved uses the question and answer tokens; consequently, there is a lot of similarity between the knowledge, question and answer tokens. This though helpful in answering also introduces noise and confusion.

Secondly, the input does not enable the comparison of different answers using attention layers. Cross-answer attention is needed to answer some questions which are comparative in nature. For example, a sample question from OpenBookQA: 

\vspace{0.5pt}
\noindent
     \fbox{\begin{minipage}{19em}
    \textbf{Question:} \textit{ Owls are likely to hunt at? } \\ a) 3:00 PM b) \textbf{2:00 AM} c) 6:00 PM d)  7:00 AM
    \end{minipage}
    }
\vspace{0.5pt}    

Entailment models may falter in such comparative questions, as they do not compare different answer options, and are only aware of one answer option at a time.

Finally, the set and order of facts retrieved for each answer is unique for an answer, but there may be sentences retrieved which are common to all the answer options. These sentences are relevant to the question, and attention with these facts with all the answer options together should enable the model to be able to discriminate between the correct and incorrect answer options. We use the above insights to develop our input and knowledge fusion modules. 

\paragraph{Input Description:}
We first categorize the retrieved facts into two classes. Let $C$ denote the set of facts which are present in the knowledge retrieved for each answer option. Let $U_{i}$ denote the set of facts which is unique to an answer option. While creating the sets we  maintain the order of the facts, after retrieval and semantic knowledge ranking. For creating $C$, we count the appearance of each sentence across different $U_{i}$ and multiply the max score for this sentence from the ranking model. We sort the sentences in decreasing order of this score. We build our inputs as follows. 

We concat the unique knowledge $U_{i}$ to the question similar to the above mentioned input to the entailment model . This input is the ``per answer option input''  defined as $\{[CLS_{A_i}] \, U_i \, Q \, [s] \, A_i \, [s] \}$.

We create another single input in which we concat the question, all the answer options and the common knowledge $C$. The ``common input'' is defined as $\{[CLS_C] \, Q\, [s]  \, A_1\, \dots [s] \, A_4 \,[s]\, C \, [s]\} $ .

For a four-way multiple choice question, we give five total inputs to BERT. Similarly for an eight-way multiple choice question, we give nine inputs.

\paragraph{Model Description:}
We feed each ``per answer input'' to BERT-based language models. We also feed the ``common input'' to the BERT-based language models. We take the corresponding embeddings of the class token, $[CLS_{A_i}]$, for each answer specific input, and concatenate the class token, $[CLS_C]$ from the ``common input'' . Let $V$ be the final concatenated vector.
\begin{equation*}
    \begin{split}
        & CLS_{A_i} = \beta(CLS,U_i,Q,A_i)  \\
        & CLS_{C} = \beta(CLS,C,Q,A_{1..4}) \\
        & V = [CLS_{A_i}:CLS_{C}]
    \end{split}
\end{equation*}
 
We feed this knowledge enriched vector $V$ through a two-layer feedforward network with an intermediate GeLU activation \cite{gelu} and BertLayerNorm \cite{devlin2019bert}. We have a final softmax layer to get probabilities.
\begin{equation*}
    \begin{split}
        & logits_i = FF2(BNorm(GeLU(FF1(V)))) \\
        & score(Q,A_i,U_i,C) = softmax(logits_i) \\
    \end{split}
\end{equation*}
where, $FF1$ and $FF2$ are feedforward networks, $BNorm$ is the BertLayerNorm and $GeLU$ is the GeLU unit. We train the model using cross-entropy loss between predictions and gold answer labels. \footnote{More training details is in Supplementary Materials.}

\begin{table*}[t]
\centering
\resizebox{\linewidth}{!}{
\begin{tabular}{ccc|c|c|c|c|c|c}
\hline
 &  &  & \multicolumn{3}{c}{Step 1 of Retrieval (\%) $\uparrow$ } & \multicolumn{3}{c}{Step 2 of Retrieval (\%) $\uparrow$ } \\ \hline
Model & Accuracy \% $\uparrow$ & Dataset & F1\: F2 : R@5 & F1\: F2 : R@10  & F1 \& F2 : R@10  & F1\: F2 : R@5 & F1\: F2 : R@10  & F1 \& F2 : R@10 \\ \hline
\multirow{2}{*}{Simple IR} & \multirow{2}{*}{N/A} & OpenBook & 33.50\: N/A & 42.60\: N/A & N/A & N/A\: N/A & N/A\: N/A & N/A \\ \cline{3-9} 
 &  & QASC & 34.32\: 11.45 & 47.54\: 14.78 & 08.12 & 33.18\: 18.78 & 47.78\: 24.56 & 11.34 \\ \hline
\multirow{2}{*}{BERT} & \multirow{2}{*}{88.32} & OpenBook & 54.60\: N/A & 65.80\: N/A & N/A & N/A\: N/A & N/A\: N/A & N/A \\ \cline{3-9} 
 &  & QASC & 46.80, 22.50 & 51.80\: 29.67 & 14.44 & 48.60\: 27.85 & 50.30\: 29.33 & 15.88 \\ \hline
\multirow{2}{*}{RoBERTa} & \multirow{2}{*}{91.56} & OpenBook & \textbf{59.62}\: N/A & \textbf{79.60}\: N/A & N/A & N/A\: N/A & N/A\: N/A & N/A \\ \cline{3-9} 
 &  & QASC & 49.32\: 28.38 & 55.80\: \underline{31.35} & 15.56 & \textbf{51.40}\: \textbf{32.56} & \textbf{57.68}\: \textbf{35.35} & \textbf{19.75} \\ \hline
\end{tabular}
}
\caption{Results for Semantic Knowledge Ranking. $F_1$ and $F_2$ represent the two core knowledge facts. Accuracy is the classification accuracy of the classifiers on validation set. Recall@N (R@N) is the measure of the fact being present in the top N retrieved sentences. $F_1$ \& $F_2$ represent both the facts are present in the top 10. For OpenBookQA we do not have annotations for gold $F_2$. Best scores are marked in Bold. }
\label{tab:skr}
\end{table*}


\section{Experimental Results and Analysis}
For Semantic Knowledge Ranking, we compare with a baseline of simple Information Retrieval using Elasticsearch. For Question Answering, we compare with a baseline of RoBERTa with and without semantically ranked knowledge. Each model is trained with a hyperparameter budget of ten runs \cite{dodge-etal-2019-show}, and the mean of the accuracies are reported. 

\begin{table}[]
\centering
\small
\begin{tabular}{l|l}
\hline
\multicolumn{2}{l}{Base Model: RoBERTa + RACE} \\ \hline \hline
Module                   & Dev Acc ($\Delta$) \% $\uparrow$ \\ \hline
No Knowledge             & 59.40               \\ \hline
Step 1                   & 62.40 \: (+3.0)        \\ \hline
Step 1 + SKR              & 66.70 \: (+4.3)        \\ \hline
Step 1 + KF             & 70.50 \: (+3.8)        \\ \hline
Step 1 + KF + SKR        & 74.60 \: (+4.1)        \\ \hline
Step 2                   & 82.50 \: (+7.9)        \\ \hline
Step 2 + SKR             & 83.90 \: (+1.4)        \\ \hline 
Step 2 + KF              & 84.20 \: (+0.3)        \\ \hline
Step 2 + KF + SKR        & 85.20 \: (+1.0)        \\ \hline
\end{tabular}
\caption{Ablation studies on the QASC dataset.  Step 1 and 2 correspond to different steps of Multi-step Knowledge Retrieval. SKR is semantic knowledge ranking. KF is our knowledge fusion model. 900 samples in the Validation set.}
\label{tab:ablation}
\end{table}

\paragraph{Semantic Knowledge Ranking:}
Table \ref{tab:skr} shows the accuracy of the BERT-based semantic knowledge ranking model. Table \ref{tab:skr} shows the impact of knowledge ranking on the precision of retrieval of gold annotated knowledge facts for OpenBookQA and QASC validation set. We compare the impact of knowledge ranking in both the steps of Multi-Step Knowledge Retrieval. We can observe that Semantic Knowledge Ranking considerably improves the precision, notably P@10 of $F_2$ facts for single-step retrieval. 

\paragraph{Ablation Studies:}
Table \ref{tab:ablation} shows the impact of each of our components on the accuracy in the QASC dataset. We add our modules over the base model of RoBERTA pre-trained on  RACE \cite{lai2017race}.
We observe that the task of QASC needs external knowledge, as the accuracy of no knowledge model is relatively low. Each of our modules, contribute to the overall increase in accuracy. Semantic knowledge ranking improves the accuracy of Step 1 by a large margin. So does the multi-step knowledge retrieval. Both the techniques have the same effect of increase in P@10 of $F_1$ and $F_2$, bringing more relevant facts for the model to answer correctly. Our Knowledge Fusion Model further improves accuracy, showing we can utilize knowledge more effectively. The increase in accuracy from knowledge fusion is especially significant using the facts retrieved from Step 1. 

\paragraph{OpenBookQA and QASC Results:}
\begin{table}[]
\centering
\small
\begin{tabular}{c|c}
\hline
Model                                      &  Acc \% $\uparrow$ \\ \hline \hline
Human & 91.7          \\ \hline
\citeauthor{Sun2018ImprovingMR} & 56.0          \\ \hline
Microsoft BERT MT                          & 64.0          \\ \hline
\citeauthor{pan2019improving}            & 70.0          \\ \hline
AristoBERTv7                               & 72.0          \\ \hline
\citeauthor{banerjee-etal-2019-careful}          & 72.0          \\ \hline
\citeauthor{AristoRoBERTaV7}                            & 77.8          \\ \hline 
RoBERTa + Step2                            & 76.4          \\ \hline \hline
Ours: Step2+SKR+KF                         & \textbf{80.0}          \\ \hline
\end{tabular}

\caption{OpenBookQA test set comparison of different models. The current state-of-the-art is AllenAI's AristoRoBERTaV7, which is a no knowledge RoBERTA model pre-trained on multiple datasets. Our model is with semantic knowledge ranking \& knowledge fusion.}
\label{tab:obqares}
\end{table}

\begin{figure}[t]
\begin{center}
  \includegraphics[width=0.9\linewidth]{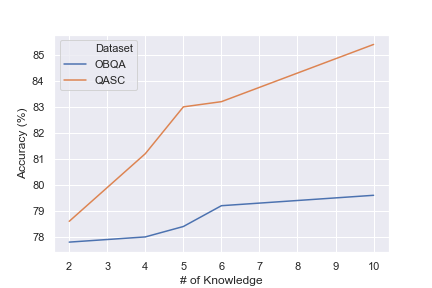}
\end{center}
   \caption{Impact of knowledge on the question answering task on the respective validation sets. More than 10 is limited by BERT max token length.}
\label{fig:ir}
\end{figure}

Table \ref{tab:obqares} and \ref{tab:qascres} compares our best model to the previous work on OpenBookQA and QASC. For QASC, we compare with our stronger RoBERTa baselines. Figure \ref{fig:ir}, shows the validation accuracy versus the number of facts retrieved. As we can see, increasing the facts increases the accuracy as more appropriate facts are seen by the model.

\begin{table}[]
\centering
\resizebox{\linewidth}{!}{
\begin{tabular}{lc|c|c}
\hline
                             & Model                      & Dev Accuracy & Test Accuracy  \\ \hline
                             \hline
\multirow{3}{*}{\citeauthor{khot2019qasc}} 
& Human          & -        & 93.00    \\ \cline{2-4}
& Random          & 12.50        & 12.50   \\ \cline{2-4}
& BERT+ Step 1               & 69.50        & 62.60          \\ \cline{2-4} 
& BERT+ Step 2               & 72.90        & 68.48          \\ \cline{2-4} 
& BERTwwm + Step 2           & 78.00        & {\ul 73.15}    \\ 

                             \hline
                             \hline
\multirow{4}{*}{Ours}        & RoBERTA+ Step 1 + KF + SKR & 74.60        &  68.80          \\ \cline{2-4} 
                             & RoBERTA+ Step 2            & 82.50        &  79.28            \\ \cline{2-4} 
                             & RoBERTA+ Step 2 + KF       & 84.20        &  80.11          \\ \cline{2-4} 
                             & RoBERTA+ Step 2 + SKR + KF & 85.20        & \textbf{80.43} \\ \hline
\end{tabular}
}
\caption{QASC validation and test set comparison between the baselines, the proposed models in \citeauthor{khot2019qasc} and our models. We compare our models with a stronger baseline of RoBERTA with knowledge. The best score is marked in bold. Previous best is underlined. BERT : BERT Large, BERTwwm : BERT Large whole word masked, RoBERTA : RoBERTA large. }
\label{tab:qascres}
\end{table}

\begin{figure}[t]
\begin{center}
  \includegraphics[width=\linewidth]{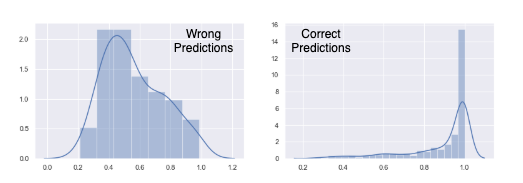}
\end{center}
   \caption{Distribution of prediction confidence of the Best Model for QASC Validation set.}
\label{fig:conf}
\end{figure}

\paragraph{Model Analysis:}
Figure \ref{fig:conf} shows our model ismore  confident  when  it  predicts  the  correct  an-swer compared to when it predicts the wrong answer. QASC has eight answer options, which increases the presence of distracting facts, and confusing answer options. Since our models use attention, and use statistical correlation, even though we retrieve relevant facts, the model predicts the answers which have the highest correlation with the facts. Our approach is to improve semantic ranking and knowledge composition to push the quality of knowledge in each step, which leads improving accuracy. Although using attention-based models for ranking brings facts that attend to all question-answer pairs, our knowledge fusion model is able to understand the appropriate knowledge. Our analysis shows that, the input creation algorithm for knowledge fusion model acts as another source for knowledge ranking, and the ``common input'' enables the model to distinguish between answers.

\paragraph{Errors in Semantic Knowledge Retrieval:}
The semantic knowledge ranking model has comparatively high accuracy (91.56\%). The errors are mostly false positives, i.e, classification of question-wrong answer and corresponding retrieved facts using the wrong answer. These are the sources of noise for the downstream question answering task. The alternative of only using the question to rank knowledge retrieved performs even worse. This indicates semantic knowledge ranking needs question-answer pair, and is a challenging task.

\paragraph{Errors in Question Answering:}
We analyzed the 100 errors made in OpenBookQA and the 137 errors made in QASC. We can broadly classify the errors made in question answering into four categories: Answering needs Complex Reasoning, Confusing fact is Retrieved, Knowledge Retrieval Failure, and Knowledge Composition Failure. There are few examples in OpenBookQA where more complex reasoning such as Temporal, Qualitative, Conjunctive and Negation is required.
Following is an example of Conjunctive Reasoning:\footnote{Correct Answers are in Bold. Predicted Answers are in italics.}

\vspace{1pt}
\noindent
\fbox{\begin{minipage}{0.97\linewidth}
\fontmed
    \textbf{Question:} Which pair don't reproduce the same?
    \\ (A) rabbit and hare (B) \textit{mule and hinny} (C) \textbf{cat and catfish} (D) caterpillar and butterfly
\end{minipage}}\\

\noindent
\fbox{\begin{minipage}{0.97\linewidth}
\fontmed
    \textbf{Question:} Astronomy can be used for what? 
    \\(A) \textit{Communication} (B) safe operation (C) vision (D) homeostasis (E) gardening (F) cooking (G) \textbf{navigation} (H) architecture\\
    \textbf{Fact Retrieved:} what is radio astronomy. a radio is used for communication.
\end{minipage}}\\

Above is an example from QASC, where  a confusing fact is retrieved that is semantically related to the question, but  supporting the wrong answer; leading the model to perform an incorrect multi-hop reasoning.

Knowledge Retrieval Failure corresponds to 72\% of the total errors in OpenBookQA. In QASC, out of 137 errors, 52 had correct $F_1$, 40 had correct $F_2$ and 25 had both $F_1$ and $F_2$ in the top ten. These errors can be mitigated by better retrieval and composition. Improving attention to perform better context-dependent similarity should enable models to distinguish between relevant and irrelevant facts.

\section{Conclusion and Future Work}
In this work, we have pushed the current state-of-the-art by 2.2\% on OpenBookQA and 7.28\% on QASC, two tasks that need external knowledge and knowledge composition for question answering. Our Semantic Knowledge Ranking and Knowledge Fusion question answering model over the BERT-based language model demonstrably improves the performance on OpenBookQA and QASC. We also provide a dataset to learn Semantic Knowledge Ranking using the annotations present in QASC, OpenBookQA, and SciTail. We have analyzed the performance of the components in our QA system. Our analysis shows the need to further improve knowledge ranking and knowledge composition.




\bibliography{anthology,acl2020}
\bibliographystyle{acl_natbib}

\clearpage

\appendix

\section{Appendices}
\label{sec:appendix}

\section{Supplemental Material}
\label{sec:supplemental}
\subsection{Training Semantic Knowledge Ranking}
We use the HuggingFace \cite{Wolf2019HuggingFacesTS} and Pytorch Deep learning framework \cite{NEURIPS2019_9015}.
We will make our code available at Github Code Link: xx.github.com.

\paragraph{Model Training}
 We train the models with learning rates in the range [1e-6,5e-5], batch sizes of [16,32,48,64], linear weight-decay in range [0.001,0.1] and warm-up steps in range of [100,1000]. 
 
\subsection{Knowledge Fusion Question Answering}
\paragraph{Model Training}
 We train the model with following hyperparameters, learning rates in the range [1e-6,5e-5], batch sizes of [16,32,48,64], linear weight-decay in range [0.001,0.1] and warm-up steps in range of [100,1000]. 

\end{document}